\documentclass[sigconf]{acmart}
\usepackage[ruled,linesnumbered]{algorithm2e}
\usepackage{adjustbox}
\usepackage{amsmath}
\AtBeginDocument{%
  \providecommand\BibTeX{{%
    \normalfont B\kern-0.5em{\scshape i\kern-0.25em b}\kern-0.8em\TeX}}}

\copyrightyear{2023}
\acmYear{2023}
\setcopyright{acmlicensed}\acmConference[CIKM '23]{Proceedings of the 32nd ACM International Conference on Information and Knowledge Management}{October 21--25, 2023}{Birmingham, United Kingdom}
\acmBooktitle{Proceedings of the 32nd ACM International Conference on Information and Knowledge Management (CIKM '23), October 21--25, 2023, Birmingham, United Kingdom}
\acmPrice{15.00}
\acmDOI{10.1145/3583780.3615497}
\acmISBN{979-8-4007-0124-5/23/10}
\begin{document}


\title{Retention Is All You Need}



\author{Karishma Mohiuddin}
\affiliation{%
\institution{University of Bonn}
\city{Bonn}
\country{Germany}
\postcode{53113}
}\email{natz.karishma@gmail.com}

\author{Mirza Ariful Alam}
\affiliation{%
\institution{University of Rajshahi}
\city{Rajshahi}
\country{Bangladesh}}
\email{mirza.hridoy@gmail.com}

\author{Mirza Mohtashim Alam}
\affiliation{%
\institution{FIZ Karlsruhe}
\city{Karlsruhe}
\country{Germany}
}\email{turzo.mohtasim@gmail.com}

\author{Pascal Welke}
\affiliation{%
\institution{TU Wien}
\city{Vienna}
\country{Austria}}
\email{pascal.welke@tuwien.ac.at}

\author{Michael Martin}
\affiliation{%
\institution{InfAI}
\city{Leipzig}
\country{Germany}}
\email{martin@infai.org}

\author{Jens Lehmann}
\authornote{Work done outside of Amazon.}
\affiliation{%
\institution{TU Dresden, InfAI, Amazon}
\city{Dresden}
\country{Germany}}
\email{jens.lehmann@tu-dresden.de}

\author{Sahar Vahdati}
\affiliation{%
\institution{InfAI}
\city{Dresden}
\country{Germany}}
\email{vahdati@infai.org}


\begin{abstract}
Skilled employees are the most important pillars of an organization. Despite this, most organizations face high attrition and turnover rates. While several machine learning models have been developed to analyze attrition and its causal factors, the interpretations of those models remain opaque.
In this paper, we propose the HR-DSS approach, which stands for Human Resource (HR) Decision Support System, and uses explainable AI for employee attrition problems.
The system is designed to assist HR departments 
in interpreting the predictions provided by machine learning models. 
In our experiments, we employ eight machine learning models to provide predictions.
We further process the results achieved by the best-performing model by the SHAP explainability process and use the SHAP values to generate natural language explanations which can be valuable for HR. 
Furthermore, using ``What-if-analysis'', we aim to observe plausible causes for attrition of an individual employee. 
The results show that by adjusting the specific dominant features of each individual, employee attrition can turn into employee retention through informative business decisions. 
\end{abstract}





\maketitle

\section{Introduction}
Attrition (voluntary leave of employees) and turnover (employee replacement) are among the major challenges of any business. 
Employee retention is crucial for reducing training and recruitment costs while preserving talent and internal knowledge for any businesses~\cite{phillips2003managing}.
The job crisis sparked by the COVID-19 pandemic serves as an additional reminder on the significance of employee attrition analysis ~\cite{parker_horowitz_minkin_2022}. 
To retain employees, companies should emphasize on the causal factors such as salary, promotion, work environment, job satisfaction, and stock option beforehand.
Current machine learning techniques for analyzing attrition and its causal determinants often rely on black-box approaches, i.e. their results are not easily interpretable by human resource experts. 

We addressed this practical issue by combining simple ML models with explainable AI (XAI) techniques aiming to reduce barriers for adoption in practice.
During the data preparation phase, we detected and removed outliers, fixed class imbalance and added weights on specific features.
In the data analysis phase, we conducted a predictive analysis using eight Machine Learning (ML) models.
Finally, we employed the XAI library SHAP (SHapley Additive exPlanations) ~\cite{NIPS2017_7062,lundberg2020local2global,lundberg2018explainable} on the ML model that achieves the highest accuracy. 
Further, for the output predictions provided by the best model (in our case XGB), we gathered the correctly predicted instances and incorrectly predicted instances. 
Since we are focusing on the application for human resources (HR), we have integrated a natural language generation module using OpenAI to output explanations in natural language. 
The whole pipeline is facilitated with an explainer dashboard to further interpret the features per each employee. 
The specific causal factors of attrition corresponding to individual employees can be understood and adjusted by ’What-if-analysis’.
The explanation provided in natural language can assist HR in a more critical decision-making process for retention policy.
\begin{figure*}[t!]
     \includegraphics[trim={1cm 1cm 2cm 1cm},scale=0.45]{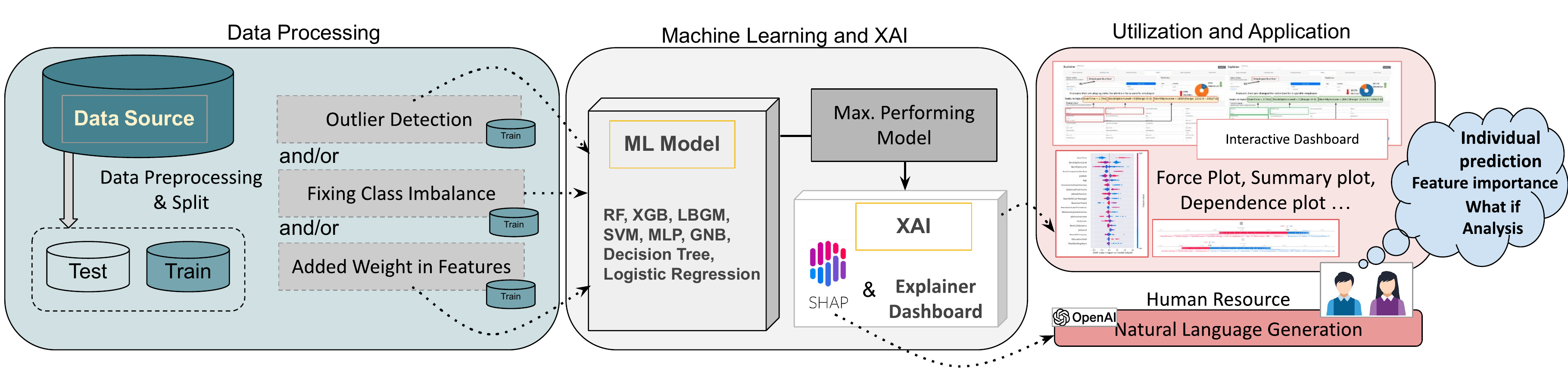}
      \caption{The workflow of the proposed decision support system (HR-DSS).}
     \label{fig:framework}
 \end{figure*}
\section{Related Work}

Different ML models have been employed to predict the importance of features in attrition~\cite{alduayj2018predicting,zhao2018employee}.
XGBoost~\cite{chen2016xgboost} is one of the predictive models that is used in several studies~\cite{ajit2016prediction,zhao2018employee}. 
In \cite{ajit2016prediction}, XGBoost outperforms other ML models for attrition prediction.
Other works use classical approaches to predict the causal factors of the attrition.
In\cite{pratt2021employee} the highest cross validation accuracy of 85.12\% is achieved by a random forest~\cite{liu2012new} among GNB, KNN, SVM, Decision Tree, Logistic Regression.
Features, such as monthly income, age, and daily rate, are suggested to be the key factors of employee turnover.
Such approaches statistically evaluate and learn from historical data, without providing any explanation for the predictions. 
SHAP (SHapley Additive exPlanations)~\cite{NIPS2017_7062,lundberg2020local2global,lundberg2018explainable} is a game theoretic approach to explain the output of any machine learning model.
SHAP is widely used in several domains including attrition analysis~\cite{sabbineni2020understanding}.
Recently, ~\cite{sekaran2022interpreting} applied the SHAP library~\cite{lundberg2017unified,sekaran2022interpreting} and LIME~\cite{dieber2020model} in order to interpret the predictions of ML models. 
However, these results did not consider the class imbalance issue typically present in employee attrition settings. 
We will further explore this in the paper.
To build a decision support system, we have employed an outlier detection and assigned weights to important features to achieve a higher accuracy of predictions. 

\section{Approach}
In this section, we present the design and development of our approach.
The proposed decision support system (DSS) is called HR-DSS and uses a SHAP library to provide explainability on top of the predictive models that are used for learning. 
In order to explain the SHAP values in natural language, we have used the OpenAI GPT-4\footnote{\url{https://openai.com/gpt-4}} completion library.
The SHAP values for the best performing ML model are constructed as prompts with the conditions and guidelines to obtain the explanation in natural language.
We deployed an XAI-based dashboard~\cite{exp_dashboard} that the HR personnel can observe, interact, and understand attrition prediction transparently and confidently. 
Figure~\ref{fig:framework} gives an overview of the proposed system. 
The interactive decision support system is based on the achieved predictions and serves as a reliable decision-support system.

We use the IBM HR Analytics Attrition Dataset\footnote{\url{https://www.kaggle.com/datasets/pavansubhasht/ibm-hr-analytics-attrition-dataset}}, which is a widely used synthetic dataset. The dataset contains 34 features related to attrition with a  sample size of 1470 records without any missing values. 
The dataset consists of both numerical and categorical data. 

\textbf{Obtaining Maximum Performing Model.}
The workflow starts with a list of machine learning models performing the prediction of prominent features causing attrition. 
As shown in Algorithm \ref{alg:algo}, let $\mathcal{MLs}$ be the collection of eight ML models $(M\_1,M\_2 ...M\_8)$ to be utilized (see Table~\ref{tab:my-tabletab2}). 
We aim at obtaining the best performing ML model $M_{pm}$ and collecting its maximum accuracy $\alpha_{MA}$. 
At the beginning of the Algorithm~\ref{alg:algo}, the data is read and pre-processed, and unnecessary columns (i.e., Over18, EmployeeCount, EmployeeNumber, StandardHours) were removed through a combination of manual and statistical verification. 
Then, the data was split into the train (80\%) and test (20\%) sets. 
Depending on the parameters, outlier detection and removal, class balancing, and feature weighting are applied to the training set (see Section~\ref{sec: work_proc}). Three distinct parameters are considered to represent these.
The feature engineering techniques are broadly discussed in Section~\ref{sec: work_proc}. 
We conduct $9$-fold cross-validation to identify the optimal model from $\mathcal{MLs}$. 
Subsequently, we set up and retrain the classifier $clf_{M_{pm}}$ employing the top-performing model $M_{pm}$ on the entire training dataset.
\begin{algorithm}[t]
\SetKwInOut{Input}{input}
\SetKwInOut{Output}{output}
 \Input{ML Models $\mathcal{MLs} = (M\_1,M\_2,...M\_8 )$}
 \Output{Max\_Accuracy $\alpha_{MA}$ Max\_perf\_model $M_{pm}$, best\_fitted\_model $clf_{M_{pm}}$}
 
 $D \gets read.data(data path)$ \\ 
 $cat\_dict, data \gets data\_process\_for\_ML (D)$\\
 $x\_train, y\_train,x\_test, y\_test \gets traintestsplit(data)$\\
\uIf{$outlier\_detect$ == $True$}{
              $(x\_train, y\_train)$ = $remove\_outlier(x\_train, y\_train)$ 
             }
\uIf{$Imblearn$ == $True$}{
              $(x\_train, y\_train)$ = $ImbLearn(x\_train, y\_train)$ 
             }
\uIf{$weighted\_feature$ == $True$}{
              $(x\_train, y\_train)$ = $added\_weighted\_feature(x\_train, y\_train)$ 
             }
$\alpha_{MA}, {M_{pm}} \gets 0, None $ \\

 \For{model in $\mathcal{MLs}$}{ 
            $clf \gets model.initialize(args)$ \\
           
            $\beta_{MA}  = crossvalscore(clf, x\_train, y\_train)$
            
            \uIf{$\mathbf\beta_{MA} > \alpha_{MA}$}{
              $\mathbf\alpha_{MA} \gets \beta_{MA}$~\\
              $\mathbf{M_{pm}} \gets model$
             }
 }
 $clf_{M_{pm}} \gets model.initialize(args, M_{pm})$ \\
 $clf_{M_{pm}}.train(x\_{train}, y\_{train})$

\caption{Obtain Maximum Performing Model}
\label{alg:algo}
\end{algorithm}

\textbf{Single Prediction Model Explainability.}
To understand the prediction of the best performing ML model, SHAP values are initialized. To interpret the model output clearly, we obtain the overall prediction (of the label Attrition) of the model on a particular instance of the test data. 
In Algorithm~\ref{alg:algo2}, the maximum performing model $\alpha_{MA}$, as provided by Algorithm~\ref{alg:algo} is taken as the input and is used to predict the label of a single example from the test set or it can be a specific employee instance. 

In this part, to analyze the single prediction, several parameters are taken as arguments. 
A particular employee from the test set or a new specific instance having the same features. 
We put the test features' specific\_employee in the visualization object $V_{a}$ of the SHAP library. 
Further, we also applied SHAP values to provide explanations in natural language. In this regard, we are using the OpenAI completion library wrapped with the model "text-davinci-003"~\cite{nay2023large}. The implementation of natural language generation is shown in Algorithm~\ref{alg: algo4}.

\begin{algorithm}[t]
\SetAlgoLined
\SetKwInOut{Input}{input}
\SetKwInOut{Output}{output}

\Input{max performing model $clf_{M_{pm}}$, feature names (\texttt{feature\_names}), prediction for the instance (\texttt{prediction}), specific\_employee}
\Output{Response in Natural Language }
shap\_values $\leftarrow$ build\_explainer\_on\_classifier($clf_{M_{pm}}$, specific\_employee)\;
\texttt{shap\_values\_current} $\gets$ $shap\_values(specific\_employee)$\;
\texttt{feature\_names} $\gets$ columns of \texttt{X\_train}\;
prompt $\gets$ build\_prompt (shap\_values\_current, the company's rules)\;
\ForEach{\texttt{feature\_name}, \texttt{shap\_value} in \texttt{(feature\_names, shap\_values\_single)}}{
  prompt $\gets$ construct(prompt, f"The SHAP value for feature '\{feature\_name\}' is \{shap\_value:.2f\}.");
}
prompt $\gets$ construct(prompt, "Please write the reason of stay or leave in bullet points?. Also provide suggestions how to retain the employee in bullet points") ;\\
\texttt{response} $\leftarrow$ OpenAI API call with \texttt{prompt}\;
Return \texttt{response}\;

\caption{Generate a natural language explanation for a single prediction}
\label{alg: algo4}
\end{algorithm}

To illustrate the dependence plot, several arguments are considered as the function parameter. In this case, we take the maximum performing model's SHAP value with index 1, test features of a specific employee, and pass it to the variable $V_{b}$ for the purpose of visualization. To initialize the explainer dashboard with the test data. The goal of using the explainer dashboard is to explain the prediction including what if analysis.


In the end, the dashboard is executed to analyze and have a proper observation of the interactive explanation.
\begin{algorithm}[t]
\SetKwInOut{Input}{input}
\SetKwInOut{Output}{output}
  \Input{$clf_{M_{pm}}$, specific\_employee}
  \Output{Force plot}
$ AnalyzeSingle\_pred(specific\_employee, clf_{M_{pm}})$ 

$V_{a} \gets shap_{fp}(specific\_employee)$ \\ 
return $V_{a}$
\caption{XAI model for single prediction}
\label{alg:algo2}
\end{algorithm}
For single prediction, the XAI algorithm is described in Algorithm~\ref{alg:algo2}.

\textbf{Global Prediction Model Explainabilty.}
In Algorithm~\ref{alg:algo3}, which is deployed to analyze the prediction of the full population, $\alpha_{MA}$, specific\_employee and x\_test are taken as the input parameters for the function. 
Indices of x\_test are obtained as mentioned above. To visualize the features of one empployee, specific\_employee is needed. $V_{a}$ is taken into consideration to be able to visualize the summary plot with SHAP. In this regard, the maximum performing model's Shap values and the feature names are provided. Moreover, to demonstrate the cumulative force plot, we take the expected values of maximum performing model from the explainer and the corresponding SHAP values for visualization. 

An explainer dashboard is used to provide further analysis.
The predicted indices of the test data, the maximum performing model $clf_{M_{pm}}$ and the categorical dictionary are provided in order to initialize the explainer object $explainer_{db}$.
Analysing the wrong predictions by the model is also possible. 
However, it is noteworthy that the use of this dashboard is not to provide predictions, but to explain possible causes for HR, after an employee leaves. 

\section{Data Processing}
\label{sec: work_proc}
We now describe the components of our data processing pipeline in more detail.
The IBM HR Analytics Attrition Dataset\footnote{\url{https://www.kaggle.com/datasets/pavansubhasht/ibm-hr-analytics-attrition-dataset}}, contains 34 features related to attrition with a sample size of 1470 records without any missing values.
To achieve good predictive performance with simple machine learning models we turn to classic data preprocessing techniques. Due to the privacy policy and unavailability of real public dataset of company, we took this synthetic data.

\begin{algorithm}[t]
\SetKwInOut{Input}{input}
\SetKwInOut{Output}{output}
  \Input{$M_{pm},specific\_employee, x\_test$}
  \Output{Summary plot \& Explainer Dashboard}
$AnalyzeFull\_pred(clf_{M_{pm}}, x\_test)$\;
$V_{a}$ $\gets$ $shap_{sp}({M_{pm}}.shap\_value[1]$, 
$features\_names)$ \\
$V_{b}$ $\gets$ $shap_{fp}(x\_test$, $clf_{M_{pm}}.explainer.expected\_value[1]$,$shap\_value[1])$ \\
$explainer_{db}$ $\gets$ $Initialize (clf_{M_{pm}}$, $x\_test$,$cat\_dict)$ \\
$explainer_{db}.run()$
\caption{XAI model for overall prediction}
\label{alg:algo3}
\end{algorithm}
\subsection{Class Imbalance for Learning Models}
The dataset has a class imbalance problem meaning that the total number of positive examples is much smaller than the number of negative examples. 
Therefore, re-sampling the data is essential to increase the importance of positive examples. 
The dataset contains 237 examples labeled ``Yes'' for Attrition and 1233 examples labeled ``No'' for Attrition. 
Undersampling of the negative examples leads to loss of instances from the data that may hold important information.
To avoid this, several oversampling techniques have been employed, namely SMOTE \citep{chawla2002smote}, ADASYN \citep{he2008adasyn}, as well as SMOTE + Tomek \citep{prati2004learning}), which is a hybrid version of SMOTE.

\begin{table}[]
\resizebox{\linewidth}{!}{%
\begin{tabular}{cccccc}
\hline
\textbf{ML Algorithm} & \textbf{\begin{tabular}[c]{@{}c@{}}Acc.\\ F1 Score\end{tabular}} & \textbf{Imbalanced} & \textbf{SMOTE} & \textbf{ADASYN} & \textbf{SMOTE+Tomek} \\ \hline
\multicolumn{1}{c|}{\textbf{Random Forest}}       & Acc.  & \colorbox{cyan!25}{87.76\%}   & 87.07\%  & \colorbox{cyan!25}{87.76\%}  & \colorbox{cyan!25}{87.76\%}  \\ \cline{2-6} 
\multicolumn{1}{c|}{}                             & F1 S. & 18.18\%  & 32.14\% & 28.00\%     & \colorbox{red!25}{35.09\%} \\ \hline
\multicolumn{1}{c|}{\textbf{Decision Tree}}       & Acc.  & 78.23\%   & \colorbox{cyan!25}{79.59\%}  & 77.89\%  & 78.23\%  \\ \cline{2-6} 
\multicolumn{1}{c|}{}                             & F1 S. & 17.95\%  & \colorbox{red!25}{33.33\%} & 30.70\% & 25.58\% \\ \hline
\multicolumn{1}{c|}{\textbf{GNB}}                 & Acc.  & \colorbox{cyan!25}{84.01\%}   & 57.48\%  & 59.18\%  & 58.16\%  \\ \cline{2-6} 
\multicolumn{1}{c|}{}                             & F1 S. & \colorbox{red!25}{47.19\%}  & 28.77\% & 29.41\% & 29.71\% \\ \hline
\multicolumn{1}{c|}{\textbf{Logistic Regression}} & Acc.  & \colorbox{cyan!25}{84.69\%}   & 64.63\%  & 60.88\%  & 64.97\%  \\ \cline{2-6} 
\multicolumn{1}{c|}{}                             & F1 S. & 4.25\% & \colorbox{red!25}{28.77\%} & 25.81\% & 27.97\% \\ \hline
\multicolumn{1}{c|}{\textbf{MLP}}                 & Acc.  & \colorbox{cyan!25}{85.03\%}   & 61.90\%  & 31.97\%  & 42.86\%  \\ \cline{2-6} 
\multicolumn{1}{c|}{}                             & F1 S. & 0.00\%      & 25.33\% & 25.37\% & \colorbox{red!25}{26.32\%} \\ \hline
\multicolumn{1}{c|}{\textbf{LGBM}}                & Acc.  & \colorbox{cyan!25}{88.44\%}   & \colorbox{cyan!25}{88.44\%}  & 87.76\%  & 87.41\%  \\ \cline{2-6} 
\multicolumn{1}{c|}{\textbf{Classifier}}          & F1 S. & 37.04\%  & \colorbox{red!25}{41.38\%} & 35.71\% & 37.29\% \\ \hline
\multicolumn{1}{c|}{\textbf{SVM}}                 & Acc.  & \colorbox{cyan!25}{86.73\%}   & \colorbox{cyan!25}{86.73\%}  & 13.27\%  & \colorbox{cyan!25}{86.73\%}  \\ \cline{2-6} 
\multicolumn{1}{c|}{}                             & F1 S. & 0.0\%       & 0.0\%      & \colorbox{red!25}{23.42\%} & 20.00\% \\ \hline
\multicolumn{1}{c|}{\textbf{XGB}}                 & Acc.  & \colorbox{cyan!25}{88.44\%}   & 85.37\%  & 87.76\%  & 86.73\%  \\ \cline{2-6} 
\multicolumn{1}{c|}{}                             & F1 S. & 35.29\%  & 29.51\% & \colorbox{red!25}{37.93\%} & 31.58\% \\ \hline
\end{tabular}%
}
\caption{Accuracy comparison among ML models trained on imbalanced data and balanced data.}
\label{tab:my-tabletab2}
\end{table}

These techniques are employed on the selected ML models, namely random forest, XGB, MLP, LGBM, decision tree, logistic regression, SVM, and Gaussian Naive Bayes. 
A fixed seed is used to avoid the problem of result deviation in the performance.
However, the accuracy of the balanced variant was then reduced for most of the ML models. 
The comparison among imbalanced and balanced variant concerning F1 scores is demonstrated in Table \ref{tab:my-tabletab2}. 
LGBM showed best performance of 88.44\% with SMOTE among all the ML models, LGBM with SMOTE also returned a high F1 score. 
\begin{figure}
    \includegraphics[scale= 0.35]{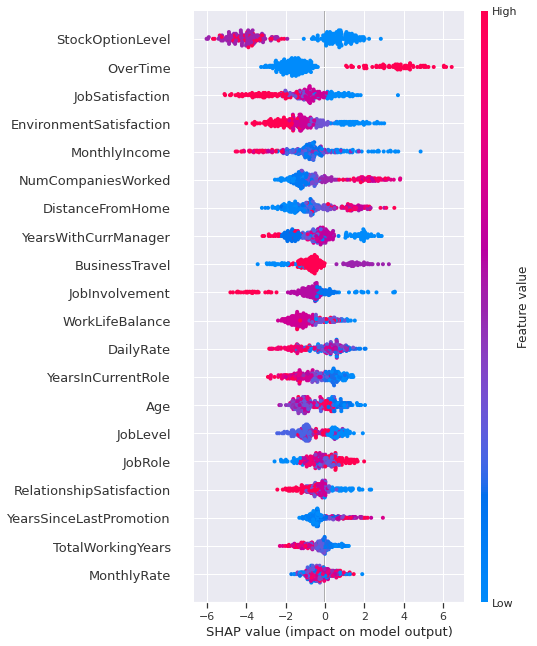}
     \caption{LGBM after balancing the data with SMOTE.}
    \label{fig:LBGM s}
\end{figure}
In Figure~\ref{fig:LBGM s}, it can be observed that there are more overlapping data points for medium (purple color) to high (red color) feature values of a particular feature. 
For example, medium to high values of StockOptionLevel are tightly mixed without being separable and many of the positive classes are shifted on the negative side in this case. 
The overlapping points of data in visualization may cause difficulties in understanding which types of employees would actually determine to leave the company. 
The best model is expected to provide the instances of positive and negative classes separated by the vertical line in the middle of the visualizations of the summary plots that divide the class labels.
Moreover, by balancing the data there are not many improvements compared to the original data.
Aligned with our observation, other works \cite{al2012using,sexton2005employee, esmaieeli2015data} also mention that accuracy measures are not reliable for imbalance classes, which can be due to severe class imbalance often mimicking a high accuracy.

\subsection{Outlier Detection and Weighted Features}
To improve the overall predictions of the ML models, an outlier detection technique has been introduced using Isolation Forest. ~\cite{liu2008isolation}
Isolation Forest has the same essence as the Decision tree algorithm. 
After providing a set of features, the algorithm randomly selects a feature from those features to remove outliers and then select a random split value to separate the maximum and minimum value of that feature. 
Besides, weights are added to StockOptionLevel and JobLevel features based on business theories \cite{oyer2005some} \cite{guan2014salary} to create a more separable group of data points because according to the theory, these two features consider as important factors of attrition. In this approach, after splitting the train and test data, Isolation forest outlier detection and weighted features are considered together to improve the performance. Additionally, the outlier detection technique and the weighted features are also experimented separately to compare the performance. According to Table \ref{tab2}, XGB has achieve improvements with improved F1 scores for every model relative to the original model. XGB with only added weights on the mentioned specific features has achieved the best result with 89.12\% accuracy. We have chosen the XGB model for interpretation with XAI for this phase to evaluate with summary plot.
\begin{figure}
    \includegraphics[scale=0.35]{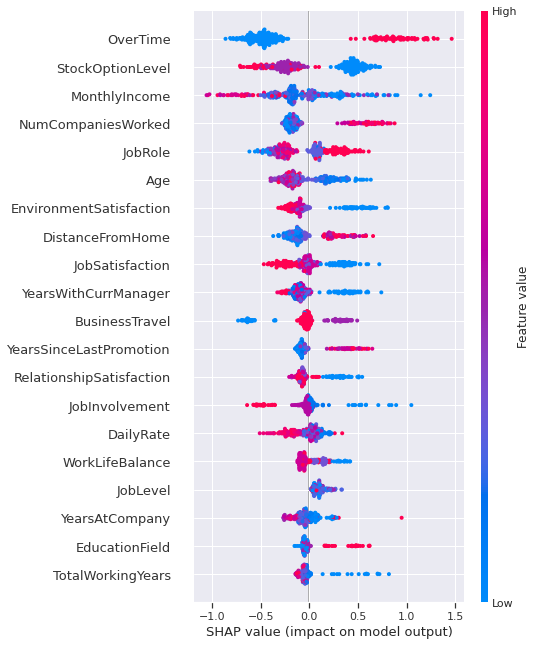}
     \caption{XGB summary plot using added weight only}
    \label{fig:XGB add}
\end{figure}

\begin{table}[]
\resizebox{\linewidth}{!}{%
\begin{tabular}{cccccc}
\hline
\textbf{ML Algorithm} &
  \textbf{\begin{tabular}[c]{@{}c@{}}Accuracy and \\ F1 Score\end{tabular}} &
  \textbf{\begin{tabular}[c]{@{}c@{}}Original\\ Data\end{tabular}} &
  \textbf{\begin{tabular}[c]{@{}c@{}}With\\ Outlier\end{tabular}} &
  \textbf{\begin{tabular}[c]{@{}c@{}}With\\ Weights\end{tabular}} &
  \textbf{\begin{tabular}[c]{@{}c@{}}With Outlier\\ and Weights\end{tabular}} \\ \hline
\multicolumn{1}{c|}{\textbf{Random Forest}}       & Acc.  & 87.76\%  & 87.76\%  & 87.07\%  & \colorbox{cyan!25}{88.44\%} \\ \cline{2-6} 
\multicolumn{1}{c|}{}                             & F1 S. & 18.18\% & 14.29\% & 17.39\% & \colorbox{red!25}{22.72\%} \\ \hline
\multicolumn{1}{c|}{\textbf{Decision Tree}}       & Acc.  & 78.23\%  & 77.89\%  & \colorbox{cyan!25}{78.57\%}  & 40.14\%  \\ \cline{2-6} 
\multicolumn{1}{c|}{}                             & F1 S. & 17.95\% & \colorbox{red!25}{34.88\%} & 22.22\% & 20.72\% \\ \hline
\multicolumn{1}{c|}{\textbf{GNB}}                 & Acc.  & \colorbox{cyan!25}{84.01\%}  & 82.31\%  & \colorbox{cyan!25}{84.01\%}  & 81.97\%  \\ \cline{2-6} 
\multicolumn{1}{c|}{}                             & F1 S. & \colorbox{red!25}{47.19\%} & \colorbox{red!25}{47.19\%} & \colorbox{red!25}{47.19\%} & 46.46\% \\ \hline
\multicolumn{1}{c|}{\textbf{Logistic Regression}} & Acc.  & 84.69\%  & 86.39\%  & 86.36\%  & \colorbox{cyan!25}{86.73\%}  \\ \cline{2-6} 
\multicolumn{1}{c|}{}                             & F1 S. & 4.26\% & 0.00\%      & 4.76\% & \colorbox{red!25}{4.88\%} \\ \hline
\multicolumn{1}{c|}{\textbf{MLP}}                 & Acc.  & 85.03\%  & 86.05\%  & 86.39\%  & \colorbox{cyan!25}{86.73\%}  \\ \cline{2-6} 
\multicolumn{1}{c|}{}                             & F1 S. & 0.00\%      & 0.00\%      & 0.00\%      & 0.00\%      \\ \hline
\multicolumn{1}{c|}{\textbf{LGBM}}                & Acc.  & \colorbox{cyan!25}{88.44\%}  & 88.10\%  & 87.07\%  & 86.39\%  \\ \cline{2-6} 
\multicolumn{1}{c|}{\textbf{Classifier}}          & F1 S. & 37.04\% & 38.60\% & \colorbox{red!25}{38.71\%} & 35.48\% \\ \hline
\multicolumn{1}{c|}{\textbf{SVM}}                 & Acc.  & \colorbox{cyan!25}{86.73\%}  & \colorbox{cyan!25}{86.73\%}  & 13.27\%  & \colorbox{cyan!25}{86.73\%}  \\ \cline{2-6} 
\multicolumn{1}{c|}{}                             & F1 S. & 0.00\%      & 0.00\%      & 0.00\%      & 0.00\%      \\ \hline
\multicolumn{1}{c|}{\textbf{XGB}}                 & Acc.  & 88.44\%  & 88.78\%  & \colorbox{cyan!25}{89.12\%}  & 88.78\%  \\ \cline{2-6} 
\multicolumn{1}{c|}{}                             & F1 S. & 35.29\% & \colorbox{red!25}{44.83\%} & 40.00\%     & 40.00\%     \\ \hline
\end{tabular}%
}
\caption{Accuracy comparison among ML algorithms trained on original data, data with added weighted features and outlier removal techniques, and a combination of both methods}
\label{tab2}
\end{table}
The accuracy of XGB improves with better F1 scores for features with added weights. Additionally, for some of the features like YearsWithCurrentManager, JobRole data points are more distinguishable in Figure~\ref{fig:XGB add}.
For XGB with weighted feature, there are less wrong predictions (32), while 262 are correctly predicted instances. Hence, XGB has been chosen for further interpretation as it is the best performance model achieving 89.12\% accuracy and its F1 Score as mentioned in Table \ref{tab2}. 
To make the explainable more understandable, NLG applied and the explainer dashboard\footnote{
\url{https://explainerdashboard.readthedocs.io/en/latest/}} has been adapted on top of SHAP to create an interactive web-based GUI platform to interpret and analyze the predictions of the model further for retention policy. 
\section{Result and Analysis} 
The XGB model has demonstrated better accuracy of 89.12\% than other ML models for this data and has faster training speed with low resources~\cite{chen2016xgboost}.
Hence, the following evaluations by SHAP visualizations are on XGB. 

\subsection{SHAP Summary Plot} 
\begin{figure}
    \centering
    \includegraphics[scale=0.35]{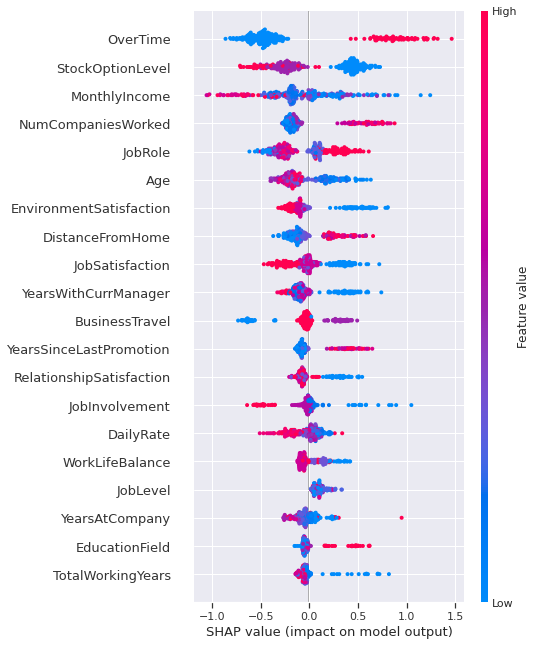}
    \caption{Summary Plot for overall prediction.}
    \label{fig:correct}
\end{figure}

The summary plot is the combination of feature importance and effect. In summary plots, the x-axis indicates the Shapley values, and
the y-axis represents the features. The feature values are arranged with color intensity between low (blue) to high (red). The summary plot in Figure \ref{fig:correct} is a global-level visualization:
Consider the 'OverTime' feature. The blue points indicate the employees do not work overtime (red dots shows the opposite). 
Hence, employees with no overtime would remain in the organization and shift towards the negative Shap value (left side), whereas the red dots mean employees with working overtime are more likely to make a positive decision towards attrition. 

\subsection{SHAP Force Plot} 
\begin{figure*}
    \centering
    \includegraphics[scale=0.50]{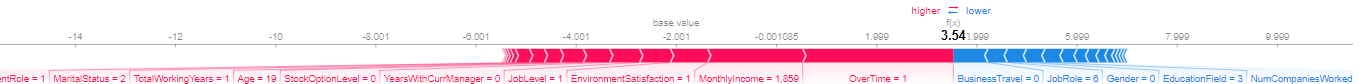} 
     \includegraphics[scale=0.53]{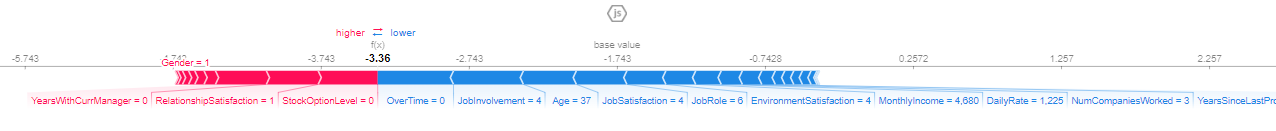}
    \caption{Force plot for individual female attrition decision (top) vs. male decision to stay (bottom) }
    \label{fig:fat}
\end{figure*}

To analyze an individual prediction, the force plot can be a useful module for understanding unique reasoning. 
For example, the upper side of Figure \ref{fig:fat} indicates that the person aged 19, a female, is strongly adamant about her attrition decision. The
red color indicates the dominant factors for the attrition where the widest bars like over time, low monthly income, age, low environmental satisfaction, no stock option plan, etc., have greater influences in her decision.
This visualization confirms that younger employees are prone to attrition than senior employees because of having moderate monthly income, working overtime and receive low stock options.~\cite{fallucchi2020predicting}\cite{srivastava2020analysis}.
On the contrary, in the bottom force plot, the SHAP output predicts and demonstrates that a male employee, aged 37, would most likely stay in the organization based on the factors of no overtime, job involvement, job satisfaction, higher monthly income, etc.

\begin{figure*}
    \centering
    \includegraphics[scale=0.25]{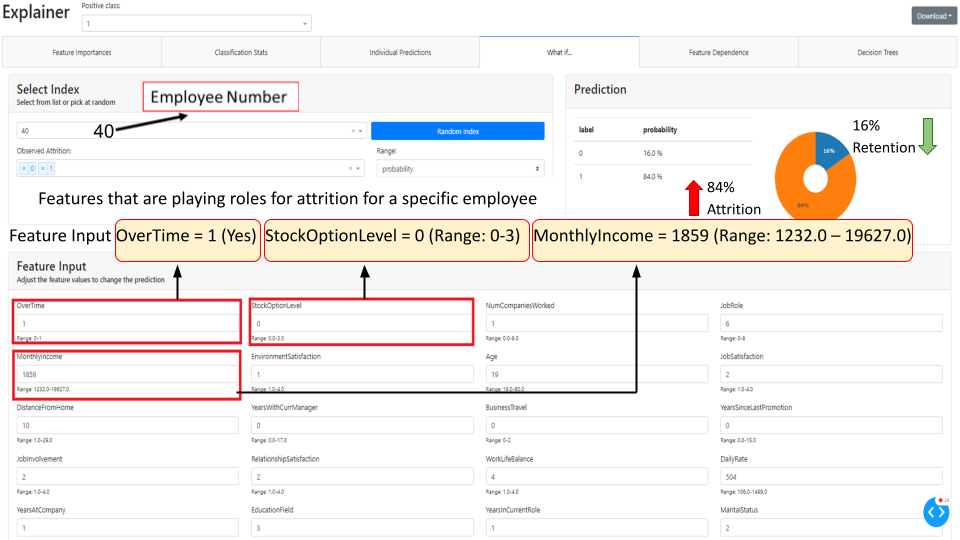} 
    \includegraphics[scale= 0.25]{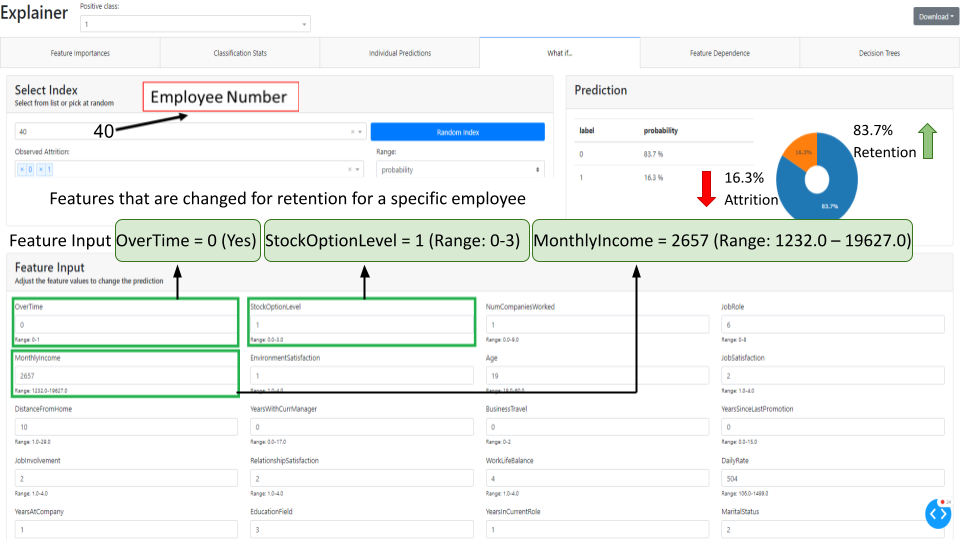} 
    \caption{Comparison between before and after using What-if-analysis for an individual employee who has probability of attrition 84\%. After using the What-if-analysis to change particular features for the employee, the attrition rate decreased to 16.3\% and the retention ratio increased to 83.7\% }
    \label{fig:exb}
\end{figure*}


\subsection{Natural Language Generation Based on SHAP Values}
We have used the Shap value output to generate natural language as described in Algorithm~\ref{alg: algo4}. The prompt for the OpenAI completion module is constructed in such a way that it contains the description what the Shap values contribute to the outcome. We have also added some dummy general policy of the company as a part of the prompt i.e, company does not offer remote work, less stock option for young employees etc. 
An example LLM output looks as follows:

Based on the SHAP values for this (woman, 19 years old) instance, the machine learning model has predicted that the employee will leave the company. Factors that contributed to this decision include: 
\begin{itemize}
\item The employee is young age (SHAP value: 0.46)
\item Their monthly income (SHAP value: 1.90)
\item Their level of stock option (SHAP value: 0.53)
\item Their years with current manager (SHAP value: 0.60)
\item Their overtime hours (SHAP value: 3.01)
\item The employees job level (SHAP value: 0.62)
\end{itemize}
In order to retain the employee, the company can:
\begin{itemize}
\item Increase the employee's salary and level of stock option
\item Increase job opportunities and promotions that are based on performance rather than years
\item Offer part-time options and flexible work hours
\item Offer more remote work options
\item Provide additional paid leaves
\item Provide more remuneration when working extra hours
\end{itemize}

\subsection{Interactive XAI Dashboard} 
To retain talented and resourceful employees,
the business community demands transparency and reliability of ML models. 
To ensure this demand, the same SHAP values are integrated to an interactive explainer dashboard.
In Figure \ref{fig:exb}, an individual employee is visualised with a predicted 86\% chance of leaving the organization with some key attributes.
From both the force plot and the interactive dashboard, it has been observed that low monthly income, overtime, and no stock option level are the most significant criteria for plausible attrition. This has been explained directly with natural language generation.
The What-if-analysis module assists HR by analyzing the probable retention of a specific employee by increasing and decreasing the causal factors and let HR to be aware of retention policy. 
With the increment of monthly salary and increased stock option level while dissolving the over time issue, the chance of attrition of this specific employee will decrease to 16.3\%. Due to the page constraint, each of the parameters effects are not shown.
\section{Conclusion}

In this work, we investigated eight machine-learning approaches for the application of HR data for attrition and retention. Among them, the best performing model is XGB with 89.12\% accuracy in providing predictions on the considered benchmark data. Further, specific features like OverTime,
and StockOptionLevel, MonthlyIncome have been determined by model as top contributing reasons to attrition. The proposed approach, named HR-DSS, is designed as an XAI-driven decision support system with an interactive user interface. 
To further enhance the decision support system, we generated natural language from the SHAP values for the explanation to human resources.  
This work shows the impact of explainable machine learning algorithms on employee attrition,retention policy and
decision-making. 
An immediate next step for this work is to include the NLG in the dashboard and enable the investigation of other neural network-based as well as deep-learning models. To overcome ethical issue and hallucination of Open-AI, we will only use the policy/guidelines documentation of an organization by applying prompt engineering. Hence, the LLM will generate results according to the XAI SHAP values and address the issue with advice from the guidelines of that organization.
The HR-DSS will be an end-to-end AI-based intelligent system aiming at assisting human resource units in different organizations to retain highly skilled employees and measure the performance of all employees with trustworthy justifications based on explainable AI.


\bibliographystyle{ACM-Reference-Format}
\bibliography{sample-base}

\pagebreak
\appendix
\renewcommand\thefigure{\thesection.\arabic{figure}}    
\setcounter{figure}{0}   

\end{document}